\documentclass[runningheads]{llncs}
\usepackage{graphicx}
\usepackage{amssymb}
\usepackage{mathrsfs}
\usepackage[linesnumbered,ruled,vlined]{algorithm2e}
\usepackage{multirow}
\usepackage{setspace}
\usepackage{color}

\begin{document}
\title{ECL: Class-Enhancement Contrastive Learning for Long-tailed Skin Lesion Classification}
\author{Yilan Zhang\inst{1} \and Jianqi Chen\inst{1} \and Ke Wang\inst{1} \and Fengying Xie\inst{1}\thanks{Corresponding author}}

\authorrunning{Zhang et al.}
\institute{Image Processing Center, School of Astronautics, Beihang University, Beijing 100191, China  \\
\email{xfy\_73.buaa.edu.cn}}
\maketitle              % typeset the header of the contribution
\begin{abstract}
Skin image datasets often suffer from imbalanced data distribution, exacerbating the difficulty of computer-aided skin disease diagnosis.  Some recent works exploit supervised contrastive learning (SCL) for this long-tailed challenge. Despite achieving significant performance, these SCL-based methods focus more on head classes, yet ignoring the utilization of information in tail classes. In this paper, we propose class-\textbf{E}nhancement \textbf{C}ontrastive \textbf{L}earning (ECL), which enriches the information of minority classes and treats different classes equally. For information enhancement, we design a hybrid-proxy model to generate class-dependent proxies and propose a cycle update strategy for parameters optimization. A balanced-hybrid-proxy loss is designed to exploit relations between samples and proxies with different classes treated equally. Taking both ``imbalanced data" and ``imbalanced diagnosis difficulty" into account, we further present a balanced-weighted cross-entropy loss following curriculum learning schedule. Experimental results on the classification of imbalanced skin lesion data have demonstrated the superiority and effectiveness of our method. The codes can be publicly available from \url{https://github.com/zylbuaa/ECL.git}.

\keywords{Contrastive learning \and Dermoscopic image\and Long-tailed classification.}
\end{abstract}
\section{Introduction}

Skin cancer is one of the most common cancers all over the world. Serious skin diseases such as melanoma can be life-threatening, making early detection and treatment essential~\cite{cassidy2022analysis}. As computer-aided diagnosis matures, recent advances with deep learning techniques such as CNNs have significantly improved the performance of skin lesion classification~\cite{esteva2017dermatologist,hasan2023survey}. However, as a data-hungry approach, CNN models require large balanced and high-quality datasets to meet the accuracy and robustness requirements in applications, which is hard to suffice due to the long-tailed occurrence of diseases in the real-world. Long-tailed problem is usually caused by the incidence rate and the difficulty of data collection. Some diseases are common while others are rare, making it difficult to collect balanced data~\cite{li2022flat}. This will cause the head classes to account for the majority of the samples and the tail classes only have small portions. Thus, existing public skin datasets usually suffer from imbalanced problems which then results in class bias of classifier, for example, poor model performance especially on tail lesion types. 

To tackle the challenge of learning unbiased classifiers with imbalanced data, many previous works focus on three main ideas, including re-sampling data~\cite{ando2017deep,pouyanfar2018dynamic}, re-weighting loss~\cite{lin2017focal,cao2019learning,yao2021single} and re-balancing training strategies~\cite{kang2019decoupling,zhou2020bbn}. Re-sampling methods over-sample tail classes or under-sample head classes, re-weighting methods adjust the weights of losses on class-level or instance-level, and re-balancing methods decouple the representation learning and classifier learning into two stages or assign the weights between features from different sampling branches~\cite{yang2022survey}. Despite the great results achieved, these methods either manually interfere with the original data distribution or improve the accuracy of minority classes at the cost of reducing that of majority classes~\cite{lango2022makes,li2022flat}.
\begin{figure}[t]
\includegraphics[width=\textwidth]{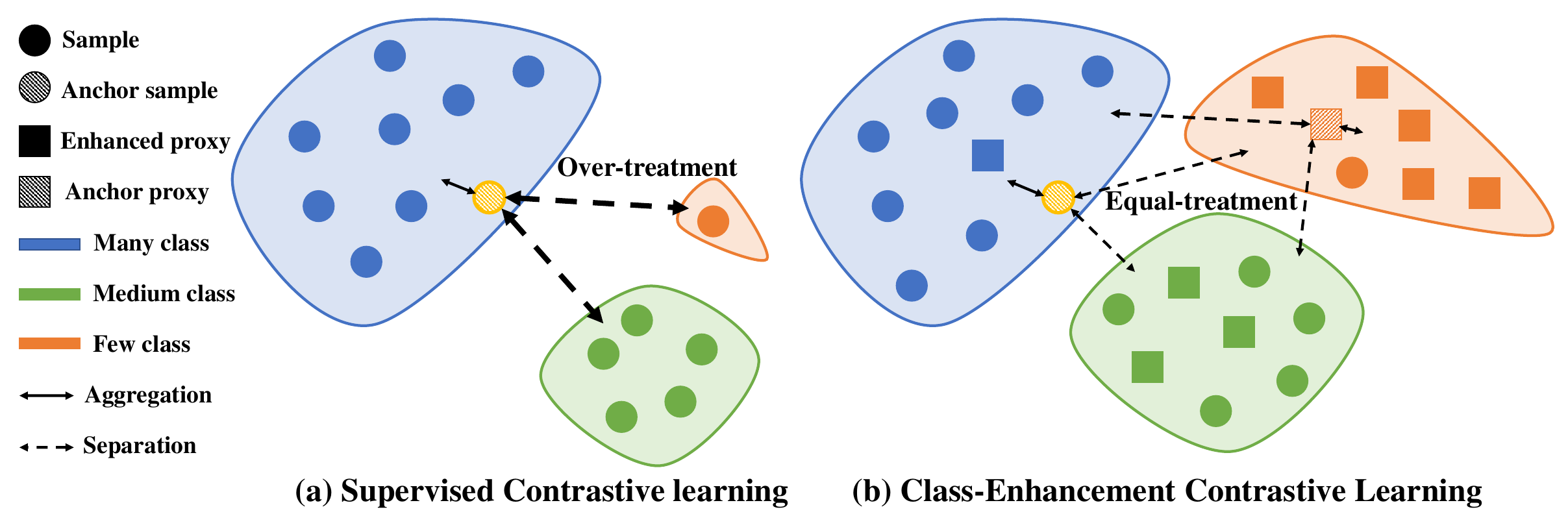}
\caption{Comparison between SCL (a) and ECL (b). In SCL, head classes are over-treated leading to optimization concentrating on head classes. By contrast, ECL utilizes the proxies to enhance the learning of tail classes and treats all classes equally according to balanced contrastive theory \cite{zhu2022balanced}. Moreover, the enriched relations in samples and proxies are helped for better representations.} \label{fig1}
\end{figure}

Recently, contrastive learning (CL) methods pose great potential for representation learning when trained on imbalanced data~\cite{li2022targeted}. Among them, supervised contrastive learning (SCL)~\cite{khosla2020supervised} aggregates semantically similar samples and separates different classes by training in pairs, leading to impressive success in long-tailed classification of both natural and medical images~\cite{marrakchi2021fighting}. However, there still remain some defects: (1) Current SCL-based methods utilize the information of minority classes insufficiently. Since tail classes are sampled with low probability, each training mini-batch inherits the long-tail distribution, making parameter updates less dependent on tail classes. (2) SCL loss focuses more on optimizing the head classes with much larger gradients than tail classes, which means tail classes are all pushed farther away from heads~\cite{zhu2022balanced}. (3) Most methods only consider the impact of sample size (``imbalanced data") on the classification accuracy of skin diseases, while ignoring the diagnostic difficulty of the diseases themselves (``imbalanced diagnosis difficulty").

To address the above issues, we propose a class-Enhancement Contrastive Learning method (ECL) for skin lesion classification, differences between SCL and ECL are illustrated in Fig.\ref{fig1}. For sufficiently utilizing the tail data information, we attempt to address the solution from a proxy-based perspective. A  proxy  can  be  regarded  as  the representative of a specific class set as learnable parameters. We propose a novel hybrid-proxy model to generate proxies for enhancing different classes with a reversed imbalanced strategy , \textit{i.e.}, the fewer samples in a class, the more proxies the class has. These learnable proxies are optimized with a cycle update strategy that captures original data distribution to mitigate the quality degradation caused by the lack of minority samples in a mini-batch. Furthermore, we propose a balanced-hybrid-proxy loss, besides introducing balanced contrastive learning (BCL)~\cite{zhu2022balanced}. The new loss treats all classes equally and utilizes sample-to-sample, proxy-to-sample and proxy-to-proxy relations to improve representation learning. Moreover, we design a balanced-weighted cross-entropy loss which follows a curriculum learning schedule by considering both imbalanced data and diagnosis difficulty.

Our contributions can be summarized as follows: (1) We propose an ECL framework for long-tailed skin lesion classification. Information of classes are enhanced by the designed hybrid-proxy model with a cycle update strategy. (2) We present a balanced-hybrid-proxy loss to balance the optimization of each class and leverage relations among samples and proxies. (3) A new balanced-weighted cross-entropy loss is designed for an unbiased classifier, which considers both ``imbalanced data" and ``imbalanced diagnosis difficulty". (4) Experimental results demonstrate that the proposed framework outperforms other state-of-the-art methods on two imbalanced dermoscopic image datasets and the ablation study shows the effectiveness of each element. 

\section{Methods}
\begin{figure}
\includegraphics[width=\textwidth]{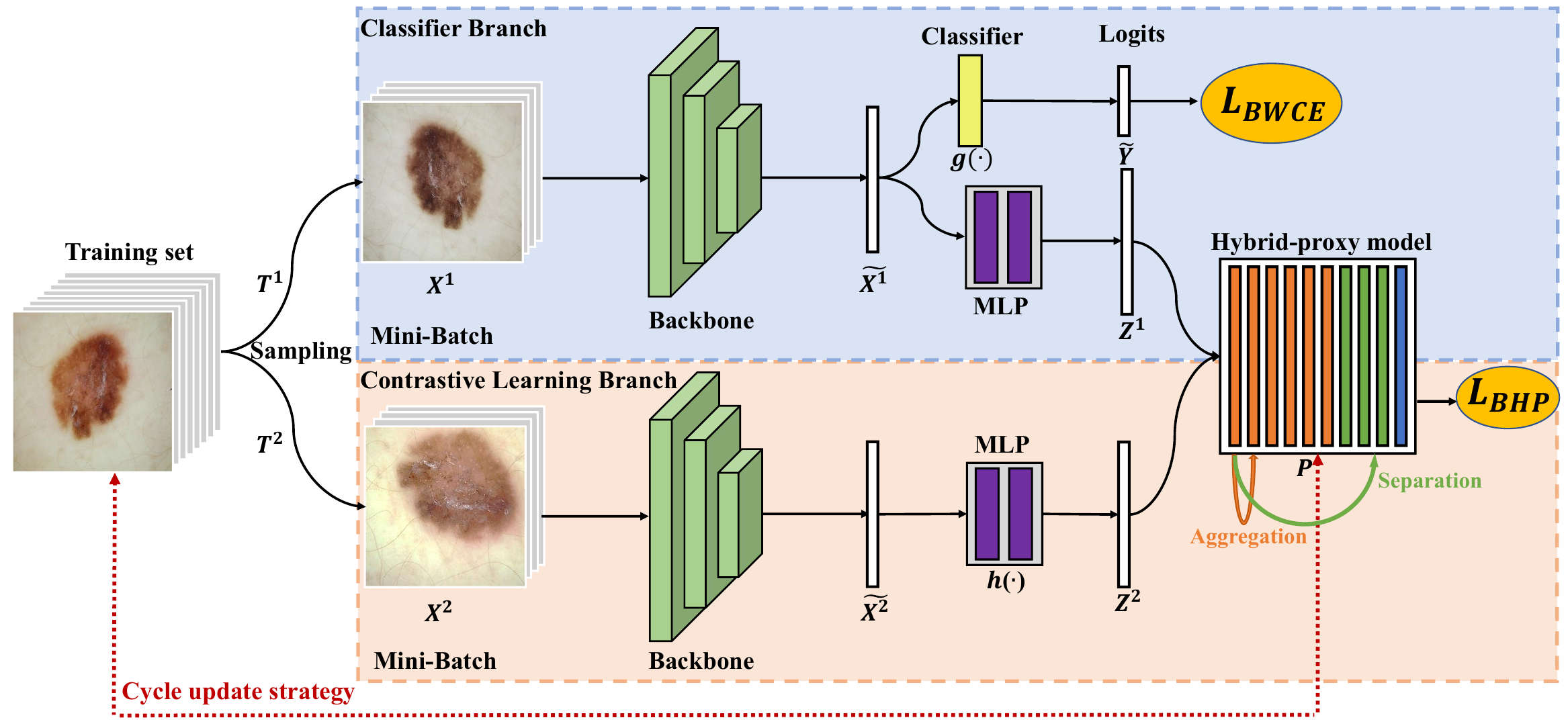}
\caption{Overall framework of the proposed ECL. ECL has two branches for classifier learning (guided by balanced-weighted cross-entropy loss $L_{BWCE}$) and contrastive learning (guided by balanced-hybrid-proxy loss $L_{BHP}$). Proxies in hybrid-proxy model are generated by a reserve imbalanced way (see Section 2.1) to strengthen the information of minority classes in a mini-batch.} \label{fig2}
\end{figure}

The overall end-to-end framework of ECL is presented in Fig. \ref{fig2}. The network consists of two parallel branches: a contrastive learning (CL) branch for representative learning and a classifier learning branch. The two branches take in different augmentations $T^{i}, i \in \left \{1,2 \right\}$ from input images $X$ and the backbone is shared between branches to learn the features $\tilde{X}^{i}, i \in \left \{1,2 \right\}$.  We use a fully connected layer as a logistic projection for classification $g(\cdot): \tilde{\mathcal{X}} \rightarrow \tilde{\mathcal{Y}}$ and a one-hidden layer MLP $h(\cdot): \tilde{\mathcal{X}} \rightarrow \mathcal{Z} \in \mathbb{R}^{d} $ as a sample embedding head where $d$ denotes the dimension. $\mathcal{L}_{2}$-normalization is applied to $\mathcal{Z}$ by using inner product as distance measurement in CL. Both the class-dependent proxies generated by hybrid-proxy model and the embeddings of samples are used to calculate balanced-weighted cross-entropy loss, thus capturing the rich relations of samples and proxies. For better representation, we design a cycle update strategy to optimize the proxies' parameters in hybrid-proxy model, together with a curriculum learning schedule for achieving unbiased classifiers. The details are introduced as follows.

\subsection{Hybrid-Proxy Model}

The proposed hybrid-proxy model consists of a set of class-dependent proxies $\mathcal{P}= \left\{ p^{c}_{k}|k\in \left\{ 1,2,...,N^{p}_{c} \right\} \right.$, $c\left. \in \left\{1,2,...,C\right\} \right\}$, $C$ is the class number, $p^{c}_{k} \in \mathbb{R}^{d}$ is the $k$-th proxy vector of class $c$, and $N^{p}_{c}$ is the proxy number in this class. Since samples in a mini-batch follow imbalanced data distribution, these proxies are designed to be generated in a reversed imbalanced way by giving more representative proxies of tail classes for enhancing the information of minority samples. Let us denote the sample number of class $c$ as $N_{c}$ and the maximum in all classes as $N_{max}$. The proxy number $N^{p}_{c}$ can be obtained by calculating the imbalanced factor $\frac{N_{max}}{N_{c}}$ of each class:
\begin{equation}
N^{p}_{c} =
\left\{ \begin{array}{ll} 
1 & N_c = N_{max} \\
\lfloor \frac{N_{max}}{10 N_{c}} \rfloor + 2 & N_c \neq N_{max} \\
\end{array} \right. 
\end{equation}
In this way, the tail classes have more proxies while head classes have less, thus alleviating the imbalanced problem in a mini-batch. 

As we know, a gradient descent algorithm will generally be executed to update the parameters after training a mini-batch of samples. However, when dealing with an imbalanced dataset, tail samples in a batch contribute little to the update of their corresponding proxies due to the low probability of being sampled. So how to get better representative proxies? Here we propose a cycle update strategy for the optimization of the parameters. Specifically, we introduce the gradient accumulation method into the training process to update proxies asynchronously. The proxies are updated only after a finished epoch that all data has been processed by the framework with the gradients accumulated. With such a strategy, tail proxies can be optimized in a view of whole data distribution, thus playing better roles in class information enhancement. Algorithm~\ref{algorithm1} presents the details of the training process.

\begin{algorithm}[t]
\small
% \setstretch{0.5}
\caption{Training process of ECL.}
\label{algorithm1}
% \SetAlgoLined
\KwIn{Training set $X$, validation set $X_{val}$, training epochs $E$, iterations $T$, batch size $B$, learning rate $lr$, stages in balanced-weighted cross-entropy loss $E_2$}
Initialize $model$ parameters $\theta$ and hybrid-proxy model $\mathcal{P}$ parameters $\phi$ \\
% \KwOut{$\tilde{\mathcal{Y}}$, labels of samples}
\For{e $in$ E}{
\For{t $in$ T}{
    {Getting a batch of samples $\left\{ x^{(1,2)}_{i},y_{i}\right\}_{B}$}
    {$\left\{ z^{(1,2)}_{i} \right\}_{B}, \left\{\tilde{y_{i}}\right\}_{B} = model (\left\{x^{(1,2)}_{i}\right\}_{B})$} \\
    \tcp{curriculum learning}
    \eIf{$e>E_2$ }{
    {$Loss(\theta,\phi) = \lambda L_{BHP}(\left\{ z^{(1,2)}_{i} \right\}_{B}, \mathcal{P}) + \mu L_{BWCE}(\left\{ y_{i},\tilde{y_{i}}\right\}_{B}, f^{e}$)}
    }{{$Loss(\theta,\phi) = \lambda L_{BHP}(\left\{ z^{(1,2)}_{i} \right\}_{B}, \mathcal{P}) + \mu L_{BWCE}(\left\{ y_{i},\tilde{y_{i}}\right\}_{B}$)}}
    {$grad^{t}_{\theta} = \nabla_{\theta}Loss(\theta), grad^{t}_{\phi} = \nabla_{\phi}Loss(\phi)$ \tcp{calculate gradients}}
    {$\theta \gets \theta - lr*grad^{t}_{\theta} $\tcp{update parameters $\theta$\ of $model$}}
}
{$\phi \gets \phi-\sum_{t}^{T} lr*grad^{t}_{\phi} $}
\tcp{update parameters $\phi$\ of $\mathcal{P}$}
\If{$e>E_2$}{$f^{e} = Validate(model, X_{val})$}}
\end{algorithm}

\subsection{Balanced-Hybrid-Proxy Loss}
To tackle the problem that SCL loss pays more attention on head classes, we introduce BCL and propose balanced-hybrid-proxy loss to treat classes equally. Given a batch of samples $\mathcal{B} = \left\{ (x^{(1,2)}_{i},y_{i})\right\}_{B}$, let $\mathcal{Z}=\left\{ z^{(1,2)}_{i} \right\}_{B} = \left \{ z^{1}_{1},z^{2}_{2},...,z^{1}_{B},z^{2}_{B} \right\}$ be the feature embeddings in a batch and $B$ denotes the batch size. For an anchor sample $z_{i} \in \mathcal{Z}$ in class $c$, we unify the positive image set as $z^{+}=\left\{ z_{j}|y_{j} = y_{i} = c, j \neq i \right\}$. Also for an anchor proxy $p^{c}_{i}$, we unify all positive proxies as $p^{+}$. The proposed balanced-hybrid-proxy loss pulls points (both samples and proxies) in the same class together, while pushes apart samples from different classes in embedding space by using dot product as a similarity measure, which can be formulated as follows:

% So the BHP loss can be merged into:
\begin{equation}
L_{BHP} = -\frac{1}{2B+\sum_{c \in C} N^{p}_{c}} \sum_{s_{i} \in \left \{ \mathcal{Z} \cup \mathcal{P} \right \} }\frac{1}{2B_{c}+N^{p}_{c}-1}\sum_{s_{j}\in \left\{z^{+} \cup p^{+} \right\}}log \frac{ exp(s_{i}\cdot s_{j}/\tau)}{E}
\end{equation}

\begin{equation}
E = \sum_{c\in C} \frac{1}{2B_{c}+N^{p}_{c}-1} \sum_{s_{k}\in \left\{\mathcal{Z}_{c} \cup \mathcal{P}_{c}\right\}}exp (s_{i} \cdot s_{k}/\tau)
\end{equation}
where $B_{c}$ means the sample number of class $c$ in a batch, $\tau$ is the temperature parameter. In addition, we further define $\mathcal{Z}_{c}$ and $\mathcal{P}_{c}$ as a subset with the label $c$ of $\mathcal{Z}$ and $\mathcal{P}$ respectively. The average operation in the denominator of balanced-hybrid-proxy loss can effectively reduce the gradients of the head classes, making an equal contribution to optimizing each class. Note that our loss differs from BCL as we enrich the learning of relations between samples and proxies. Sample-to-sample, proxy-to-sample and proxy-to-proxy relations in the proposed loss have the potential to promote network's representation learning. Moreover, as the skin datasets are often small, richer relations can effectively help form a high-quality distribution in the embedding space and improve the separation of features.

\subsection{Balanced-Weighted Cross-Entropy Loss}
Taking both ``imbalanced data" and ``imbalanced diagnosis difficulty" into consideration, we design a curriculum schedule and propose balanced-weighted cross-entropy loss to train an unbiased classifier. The training phase are divided into three stages. We first train a general classifier, then in the second stage we assign larger weight to tail classes for ``imbalanced data". In the last stage, we utilize the results on the validation set as the diagnosis difficulty indicator of skin disease types to update the weights for ``imbalanced diagnosis difficulty". The loss is given by: 

\begin{equation}
L_{BWCE} = - \frac{1}{B} \sum_{i=1}^{B} w_{i} CE(\tilde{y_{i}},y_{i})
\end{equation}
% \begin{equation}
% w_{i} = \left\{ \begin{aligned} 
% 0,& e=0 I\&
% \frac{1}{N_{c}}\frac{1}{N_{c}},& e=1
% \end{aligned}\right.
% \end{equation}

\begin{equation}
w_{i} = \left\{ \begin{array}{ll} 
1& e<E_{1} \\
(\frac{C/N_{c}}{\sum_{c\in C}1/N_{c}})^{\frac{e-E_1}{E_2-E_1}}& E_1<e<E_2\\
(\frac{C/f^{e}_{c}}{\sum_{c\in C}1/f^{e}_{c}})^{\frac{e-E_2}{E-E_2}}& E_2<e<E\\
\end{array} \right. 
\end{equation}
where $w$ denotes the weight and $\tilde{y}$ denotes the network prediction. We assume $f^{e}_{c}$ is the evaluation result of class $c$ on validation set after epoch e and we use f1-score in our experiments. The network is trained for $E$ epochs, $E_1$ and $E_2$ are hyperparameters for stages. The final loss is given by $Loss = \lambda L_{BHP} + \mu L_{BWCE}$ where $\lambda$ and $\mu$ are the hyperparameters which control the impact of losses.

\section{Experiment}
\subsection{Dataset and Implementation Details}
\subsubsection{Dataset and Evaluation Metrics.}
We evaluate the ECL on two publicly available dermoscopic datasets ISIC2018\cite{codella2018skin,tschandl2018ham10000} and ISIC2019\cite{codella2018skin,combalia2019bcn20000,tschandl2018ham10000}. The 2018 dataset consists of 10015 images in 7 classes while a larger 2019 dataset provides 25331 images in 8 classes. The imbalanced factors $\alpha = \frac{N_{max}}{N_{min}}$ of the two datasets are all $>50$ (ISIC2018 58.30 and ISIC2019 53.87), which means that skin lesion classification suffers a serious imbalanced problem. We randomly divide the samples into the training, validation and test sets as 3:1:1. 

We adopt five metrics for evaluation: accuracy (Acc), average precision (Pre), average sensitivity (Sen), macro f1-score (F1) and macro area under curve (AUC). Acc and F1 are considered as the most important metrics in this task. 

 \subsubsection{Implementation Details.}
 \begin{figure}[t]
\includegraphics[width=\textwidth]{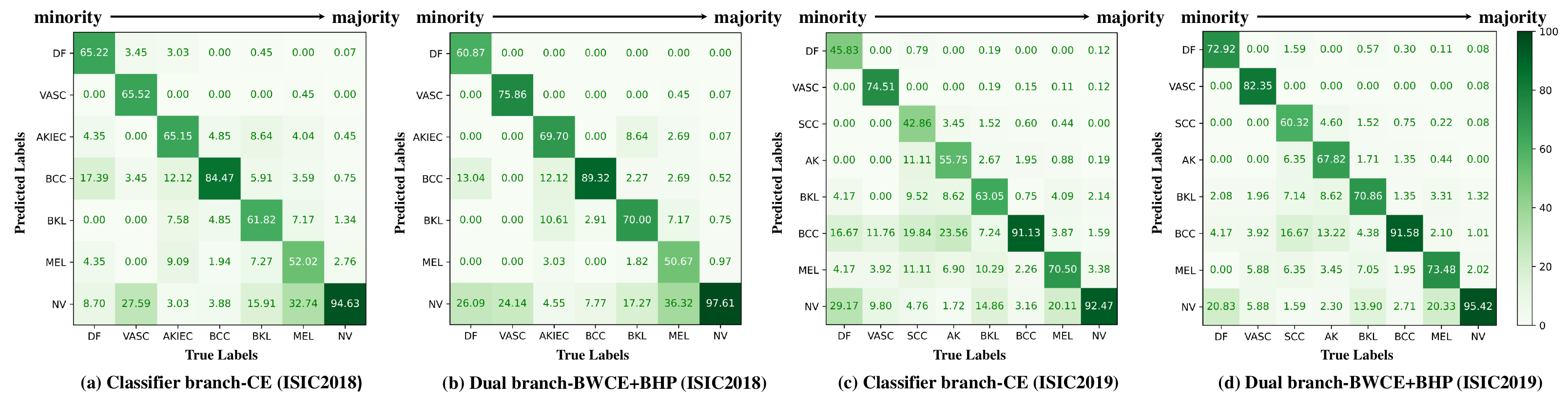}
\caption{The results of confusion matrix illustrate that ECL obtains great performance on most classes especially for minority classes.
% The numbers in matrix are the percentage of Ture labels.
} \label{fig3}
\end{figure}
The proposed algorithm is implemented in Python with Pytorch library and runs on a PC equipped with an NVIDIA A100 GPU. 
We use ResNet50~\cite{he2016deep} as backbone and the embedding dimension $d$ is set to 128. We use SGD as the optimizer with the weight decay 1e-4. The initial learning rate is set to 0.002 and decayed by cosine schedule. We train the network for 100 epochs with a batch size of 64. The hyperparameters $E_1$, $E_2$, $\tau$, $\lambda$, and $\mu$ are set to 20, 50, 0.01, 1, and 2 respectively. We use the default data augmentation strategy on ImageNet in~\cite{he2016deep} as $T_1$ for classification branch. And for CL branch, we add random grayscale, rotation, and vertical flip in $T_1$ as $T_2$ to enrich the data representations. Meanwhile, we only conduct the resize operation to ensure input size $224 \times 224 \times 3$ during testing process. 
The models with the highest Acc on validation set are chosen for testing. We conduct experiments in 3 independent runs and report the standard deviations in the supplementary material. 

\subsection{Experimental Results}

\subsubsection{Quantitative Results.}

To evaluate the performance of our ECL, we compare our method with 10 advanced methods. Among them, focal loss~\cite{lin2017focal}, LDAM-DRW~\cite{cao2019learning}, logit adjust~\cite{menon2020long}, and MWNL~\cite{yao2021single} are the re-weighting loss methods.  BBN~\cite{zhou2020bbn} is the methods based on re-balancing training strategy while Hybrid-SC~\cite{wang2021contrastive}, SCL~\cite{khosla2020supervised,marrakchi2021fighting}, BCL~\cite{zhu2022balanced}, TSC~\cite{li2022targeted} and ours are the CL-based methods. Moreover, MWNL and SCL have been verified to perform well in the skin disease classification task. To ensure fairness, we re-train all methods by rerun their released codes on our divided datasets with the same experimental settings. We also confirmed that all models have converged and choose the best eval checkpoints. The results are shown in Table~\ref{comparisom}. It can be seen that ECL has a significant advantage with the highest level in most metrics on two datasets. Noticeably, our ECL outperforms other imbalanced methods by great gains, e.g., 2.56\% in Pre on ISIC2018 compared with SCL and 4.33\% in F1 on ISIC2019 dataset compared with TSC. Furthermore, we draw the confusion matrixes after normalization in Fig. \ref{fig3}, which illustrate that ECL has significantly improved most of the categories, from minority to majority.

% \usepackage{multirow}
\begin{table}[t]
\caption{Comparison results on ISIC2018 and ISIC2019 datasets.}\label{comparisom}
\setlength\tabcolsep{2.3pt}
\renewcommand{\arraystretch}{1}
\small
\begin{tabular}{l|ccccc|ccccc}
\hline\hline
\multicolumn{1}{c|}{\multirow{2}{*}{Methods}} & \multicolumn{5}{c|}{ISIC2018}                                                      & \multicolumn{5}{c}{ISIC2019}                                                       \\
\multicolumn{1}{c|}{}                         & Acc            & Sen            & Pre            & F1             & AUC            & Acc            & Sen            & Pre            & F1             & AUC            \\ \hline
CE                                            & 83.89          & 69.56          & 73.62          & 70.34          & 94.81          & 82.41          & 67.02          & 77.32          & 70.90          & 95.37          \\
Focal Loss                                    & 84.19          & 68.78          & 76.69          & 71.38          & 94.76          & 82.05          & 64.55          & 75.93          & 68.84          & 94.82          \\
LDAM-DRW                                      & 84.20          & 71.74          & 74.65          & 71.98          & 95.22          & 82.29          & 68.08          & 74.61          & 70.84          & 95.65          \\
Logit Adjust                                  & 84.15          & 71.54          & 71.78          & 70.77          & 95.55          & 81.93          & 68.94          & 69.12          & 68.64          & 95.17          \\
MWNL                                          & 84.90          & 73.90          & 76.94          & 74.92          & \textbf{96.79} & 84.10          & 74.83          & 75.81          & 75.08          & 96.61          \\ \hline
BBN                                           & 85.57          & \textbf{74.96} & 72.40          & 72.79          & 93.72          & 83.43          & 71.78          & 78.37          & 74.42          & 95.10          \\ \hline
Hybrid-SC                                     & 86.30          & 73.93          & 75.84          & 74.34          & 96.33          & 84.69          & 70.90          & 76.87          & 73.27          & 96.67          \\
SCL                                           & 86.13          & 70.40          & 80.88          & 74.27          & 96.56          & 84.60          & 70.90          & 81.66          & 75.07          & 96.21          \\
BCL                                           & 84.92          & 72.87          & 71.15          & 71.57          & 95.61          & 83.47          & 73.52          & 74.17          & 73.50          & 95.95          \\
TSC                                           & 85.94          & 73.35          & 77.77        & 74.94          & 95.83          & 84.75          & 71.89          & 79.81          & 75.13          & 95.84          \\
Ours                                          & \textbf{87.20} & 73.01          & \textbf{83.44} & \textbf{76.76} & 96.55          & \textbf{86.11} & \textbf{76.57} & \textbf{83.22} & \textbf{79.46} & \textbf{96.78} \\ \hline\hline
\end{tabular}
\end{table}

\begin{table}
\caption{Ablation study on ISIC2019 dataset.}\label{ablation}
\setlength\tabcolsep{3pt}
\renewcommand{\arraystretch}{1}
\small
\begin{tabular}{l|l|ccccc}
\hline\hline
Methods(ISIC2019)                                                                       & Proxies             & Acc            & Sen            & Pre            & F1             & AUC            \\ \hline
Classifier branch-CE                                                                    & HPM                 & 82.41          & 67.02          & 77.32         & 70.90         & 95.37          \\
Classifier branch-BWCE                                                                  & HPM                 & 82.69          & 67.95          & 77.32          & 71.65          & 95.37          \\
Dual branch-CE+BHP                                                                      & HPM                 & 85.49          & 73.35          & 81.61          & 76.76          & 96.52 \\\hline

Dual branch-BWCE+BHP                                                                   & 2 proxies per-class & 85.52          & 74.03         & 81.46          & 77.22          & 96.53          \\
Dual branch-BWCE+BHP                                                                     & 3 proxies per-class & 85.36          & 73.49          & 83.00          & 77.53          & 96.74          \\
Dual branch-BWCE+BHP                                                                     & 4 proxies per-class & 85.79          & 74.09          & 82.03          & 77.42          & 96.53          \\\hline
\multirow{2}{*}{Dual branch-BWCE+BHP}  & w/o cycle stategy                & 85.65          & 73.48          & 83.00          & 77.40          & 96.65          \\
& HPM                 & \textbf{86.11} & \textbf{76.57} & \textbf{83.22} & \textbf{79.46} & \textbf{96.78}          \\ \hline\hline
\end{tabular}
\end{table}

\subsubsection{Ablation Study.}
To further verify the effectiveness of the designs in ECL, we conduct a detailed ablation study shown in Table~\ref{ablation} (the results on ISIC2018 are shown in supplementary material Table S2). First, we directly move the contrastive learning (CL) branch and replaced the balenced-weighted cross-entropy (BWCE) loss with cross-entropy (CE) loss. We can see from the results that adding CL branch can significantly improve the network's data representation ability with better performance than only adopting a classifier branch. And our BWCE loss can help in learning a more unbiased classifier with an improvement of 1.94\% in F1 compared with CE on ISIC2019. Then we train the ECL w/o cycle update strategy. The overall performance of the network has declined compared with training w/ the strategy, indicating that this strategy can better enhance proxies learning through the whole data distribution. In the end, we also set the proxies' number of different classes equal to explore whether the classification ability of the network is improved due to the increase in the number of proxies. With more proxies, metrics fluctuate and do not increase significantly. However, the result of using proxies generated by reversed balanced way in hybrid-proxy model (HPM) outperforms equal proxies in nearly all metrics, which proves that more proxies can effectively enhance and enrich the information of tail classes. 

\section{Conclusion}
In this work, we present a class-enhancement contrastive learning framework, named ECL, for long-tailed skin lesion classification. Hybrid-proxy model and balanced-hybrid-proxy loss are proposed to tackle the problem that SCL-based methods pay less attention to the learning of tail classes. Class-dependent proxies are generated in hybrid-proxy model to enhance information of tail classes, where rich relations between samples and proxies are utilized to improve representation learning of the network. Furthermore, blanced-weighted cross-entropy loss is designed to help train an unbiased classifier by considering both "imbalanced data" and "imbalanced diagnosis difficulty". Extensive experiments on ISIC2018 and ISIC2019 datasets have demonstrated the effectiveness and superiority of ECL over other compared methods. 

\bibliographystyle{splncs04}
\bibliography{mybibliography}
\newpage
\section*{Supplementary Material: ECL: Class-Enhancement Contrastive Learning of Long-tailed Skin Leision Classification}

\setcounter{table}{0}
% \usepackage{multirow}
\begin{table}
\renewcommand{\thetable}{S\arabic{table}}
\caption{Comparison results on ISIC2018 and ISIC2019 datasets. Data format: mean (standard deviation)}
\setlength\tabcolsep{1.2pt}
\renewcommand{\arraystretch}{1.2}
\begin{tabular}{l|ccccc}
\hline
\multicolumn{1}{c|}{\multirow{2}{*}{Methods}} & \multicolumn{5}{c}{ISIC2018}                                                                                          \\
\multicolumn{1}{c|}{}                         & Acc                   & Pre                   & Sen                   & F1                    & AUC                   \\ \hline
CE                                            & 83.89 (0.33)          & 69.56 (0.29)          & 73.62 (1.39)          & 70.34 (0.68)          & 94.81 (0.09)          \\
Focal Loss                                    & 84.19 (0.13)          & 68.78 (0.43)          & 76.69 (0.48)          & 71.38 (0.41)          & 94.76 (0.02)          \\
LDAM-DRW                                      & 84.20 (0.09)          & 71.74 (0.43)          & 74.65 (0.26)          & 71.98 (0.32)          & 95.22 (0.05)          \\
Logit Adjust                                  & 84.15 (0.27)          & 71.54 (1.30)          & 71.78 (1.10)          & 70.77 (0.04)          & 95.55 (0.04)          \\
MWNL                                          & 84.90 (0.20)          & 73.90 (0.43)          & 76.94 (0.35)          & 74.92 (0.43)          & \textbf{96.79 (0.14)} \\ \hline
BBN                                           & 85.57 (0.40)          & \textbf{74.96 (1.23)} & 72.40 (2.75)          & 72.79 (1.02)          & 93.72 (0.03)          \\ \hline
Hybrid-SC                                     & 86.30 (0.36)          & 73.93 (0.46)          & 75.84 (4.48)          & 74.34 (2.13)          & 96.33 (0.53)          \\
SCL                                           & 86.13 (0.22)          & 70.40 (0.91)          & 80.88 (0.97)          & 74.27 (0.86)          & 96.56 (0.08)          \\
BCL                                           & 84.92 (0.49)          & 72.87 (0.85)          & 71.15 (0.66)          & 71.57 (0.51)          & 95.61 (0.10)          \\
TSC                                           & 85.94 (0.46)          & 73.35 (1.80)          & 77.77 (1.60)          & 74.94 (1.66)          & 95.83 (0.19)          \\
Ours                                          & \textbf{87.20 (0.12)} & 73.01 (0.48)          & \textbf{83.44 (0.77)} & \textbf{76.76 (0.33)} & 96.55 (0.03)          \\ \hline
\multicolumn{1}{c|}{\multirow{2}{*}{Methods}} & \multicolumn{5}{c}{ISIC2019}                                                                                          \\
\multicolumn{1}{c|}{}                         & Acc                   & Pre                   & Sen                   & F1                    & AUC                   \\ \hline
CE                                            & 82.41 (0.19)          & 67.02 (0.10)          & 77.32 (0.25)          & 70.90 (0.10)          & 95.37 (0.04)          \\
Focal Loss                                    & 82.05 (0.11)          & 64.55 (0.20)          & 75.93 (0.90)          & 68.84 (0.31)          & 94.82 (0.04)          \\
LDAM-DRW                                      & 82.29 (0.10)          & 68.08 (0.34)          & 74.61 (0.12)          & 70.84 (0.23)          & 95.65 (0.02)          \\
Logit Adjust                                  & 81.93 (0.21)          & 68.94 (0.29)          & 69.12 (0.14)          & 68.64 (0.21)          & 95.17 (0.01)          \\
MWNL                                          & 84.10 (0.18)          & 74.83 (0.78)          & 75.81 (0.52)          & 75.08 (0.23)          & 96.61 (0.04)          \\ \hline
BBN                                           & 83.43 (0.10)          & 71.78 (1.32)          & 78.37 (2.18)          & 74.42 (0.33)          & 95.10 (0.07)          \\ \hline
Hybrid-SC                                     & 84.69 (0.09)          & 70.90 (0.38)          & 76.87 (0.25)          & 73.27 (0.38)          & 96.67 (0.04)          \\
SCL                                           & 84.60 (0.24)          & 70.90 (1.57)          & 81.66 (0.54)          & 75.07 (1.23)          & 96.21 (0.07)          \\
BCL                                           & 83.47 (0.10)          & 73.52 (1.40)          & 74.17 (1.12)          & 73.50 (0.29)          & 95.95 (0.03)          \\
TSC                                           & 84.75 (0.15)          & 71.89 (0.64)          & 79.81 (0.31)          & 75.13 (0.32)          & 95.84 (0.03)          \\
\textbf{Ours}                                 & \textbf{86.11 (0.16)} & \textbf{76.57 (0.94)} & \textbf{83.22 (0.10)} & \textbf{79.46 (0.58)} & \textbf{96.78 (0.09)} \\ \hline
\end{tabular}
\end{table}

\begin{table}[]
\renewcommand{\thetable}{S\arabic{table}}
\caption{Ablation study on ISIC2018 and ISIC2019 datasets. Data format: mean (standard deviation)}\label{ablation}
\renewcommand{\arraystretch}{1.2}
\begin{tabular}{l|l|ccccc}
\hline \hline 
Methods(ISIC2018)                                                                       & Proxies             & Acc                                                             & Sen                                                             & Pre                                                             & F1                                                              & AUC                                                             \\ \hline \hline
Classifier branch-CE                                                                    & HPM                 & \begin{tabular}[c]{@{}c@{}}83.89\\ (0.33)\end{tabular}          & \begin{tabular}[c]{@{}c@{}}69.56\\ (0.29)\end{tabular}          & \begin{tabular}[c]{@{}c@{}}73.62\\ (1.39)\end{tabular}          & \begin{tabular}[c]{@{}c@{}}70.34\\ (0.68)\end{tabular}          & \begin{tabular}[c]{@{}c@{}}94.81\\ (0.09)\end{tabular}          \\\hline
Classifier branch-BWCE                                                                  & HPM                 & \begin{tabular}[c]{@{}c@{}}84.83\\ (0.44)\end{tabular}          & \begin{tabular}[c]{@{}c@{}}70.13\\ (1.85)\end{tabular}          & \begin{tabular}[c]{@{}c@{}}77.38\\ (0.46)\end{tabular}          & \begin{tabular}[c]{@{}c@{}}72.28\\ (1.33)\end{tabular}          & \begin{tabular}[c]{@{}c@{}}94.94\\ (0.16)\end{tabular}          \\ \hline
Dual branch-CE+BHP                                                                      & HPM                 & \begin{tabular}[c]{@{}c@{}}86.78\\ (0.18)\end{tabular}          & \begin{tabular}[c]{@{}c@{}}72.96\\ (0.21)\end{tabular}          & \begin{tabular}[c]{@{}c@{}}81.73\\ (0.21)\end{tabular}          & \begin{tabular}[c]{@{}c@{}}76.05\\ (0.97)\end{tabular}          & \textbf{\begin{tabular}[c]{@{}c@{}}96.74\\ (0.08)\end{tabular}} \\ \hline
\begin{tabular}[c]{@{}l@{}}Dual branch-BWCE+BHP\\ w/o cycle update stategy\end{tabular} & HPM                 & \begin{tabular}[c]{@{}c@{}}86.43\\ (0.09)\end{tabular}          & \begin{tabular}[c]{@{}c@{}}72.42\\ (0.56)\end{tabular}          & \begin{tabular}[c]{@{}c@{}}81.32\\ (0.26)\end{tabular}          & \begin{tabular}[c]{@{}c@{}}75.14\\ (0.18)\end{tabular}          & \begin{tabular}[c]{@{}c@{}}96.50\\ (0.02)\end{tabular}          \\ \hline
Dual branch-BWCE+BHP                                                                    & 2 proxies per-class & \begin{tabular}[c]{@{}c@{}}86.33\\ (0.26)\end{tabular}          & \begin{tabular}[c]{@{}c@{}}71.32\\ (0.80)\end{tabular}          & \begin{tabular}[c]{@{}c@{}}81.86\\ (1.41)\end{tabular}          & \begin{tabular}[c]{@{}c@{}}75.18\\ (0.68)\end{tabular}          & \begin{tabular}[c]{@{}c@{}}96.30\\ (0.09)\end{tabular}          \\ \hline
Dual branch-BWCE+BHP                                                                    & 3 proxies per-class & \begin{tabular}[c]{@{}c@{}}86.36\\ (0.28)\end{tabular}          & \begin{tabular}[c]{@{}c@{}}70.47\\ (1.21)\end{tabular}          & \begin{tabular}[c]{@{}c@{}}82.20\\ (0.54)\end{tabular}          & \begin{tabular}[c]{@{}c@{}}74.70\\ (0.75)\end{tabular}          & \begin{tabular}[c]{@{}c@{}}96.59\\ (0.06)\end{tabular}          \\ \hline
Dual branch-BWCE+BHP                                                                    & 4 proxies per-class & \begin{tabular}[c]{@{}c@{}}86.45\\ (0.33)\end{tabular}          & \begin{tabular}[c]{@{}c@{}}71.54\\ (0.52)\end{tabular}          & \begin{tabular}[c]{@{}c@{}}80.54\\ (2.00)\end{tabular}          & \begin{tabular}[c]{@{}c@{}}75.84\\ (0.44)\end{tabular}          & \begin{tabular}[c]{@{}c@{}}96.66\\ (0.04)\end{tabular}          \\ \hline
Dual branch-BWCE+BHP                                                                    & HPM                 & \textbf{\begin{tabular}[c]{@{}c@{}}87.20\\ (0.12)\end{tabular}} & \textbf{\begin{tabular}[c]{@{}c@{}}73.02\\ (0.48)\end{tabular}} & \textbf{\begin{tabular}[c]{@{}c@{}}83.44\\ (0.77)\end{tabular}} & \textbf{\begin{tabular}[c]{@{}c@{}}76.76\\ (0.33)\end{tabular}} & \begin{tabular}[c]{@{}c@{}}96.55\\ (0.03)\end{tabular}          \\ \hline \hline
Methods(ISIC2019)                                                                       & Proxies             & Acc                                                             & Sen                                                             & Pre                                                             & F1                                                              & AUC                                                             \\ \hline \hline
Classifier branch-CE                                                                    & HPM                 & \begin{tabular}[c]{@{}c@{}}82.41\\ (0.19)\end{tabular}          & \begin{tabular}[c]{@{}c@{}}67.02\\ (0.10)\end{tabular}          & \begin{tabular}[c]{@{}c@{}}77.32\\ (0.25)\end{tabular}          & \begin{tabular}[c]{@{}c@{}}70.90\\ (0.10)\end{tabular}          & \begin{tabular}[c]{@{}c@{}}95.37\\ (0.04)\end{tabular}          \\\hline
Classifier branch-BWCE                                                                  & HPM                 & \begin{tabular}[c]{@{}c@{}}82.69\\ (0.14)\end{tabular}          & \begin{tabular}[c]{@{}c@{}}67.95\\ (0.77)\end{tabular}          & \begin{tabular}[c]{@{}c@{}}77.32\\ (0.31)\end{tabular}          & \begin{tabular}[c]{@{}c@{}}71.65\\ (0.46)\end{tabular}          & \begin{tabular}[c]{@{}c@{}}95.35\\ (0.03)\end{tabular}          \\\hline
Dual branch-CE+BHP                                                                      & HPM                 & \begin{tabular}[c]{@{}c@{}}85.49\\ (0.03)\end{tabular}          & \begin{tabular}[c]{@{}c@{}}73.35\\ (0.30)\end{tabular}          & \begin{tabular}[c]{@{}c@{}}81.61\\ (0.30)\end{tabular}          & \begin{tabular}[c]{@{}c@{}}76.76\\ (0.31)\end{tabular}          & \begin{tabular}[c]{@{}c@{}}96.52\\ (0.03)\end{tabular}          \\\hline
\begin{tabular}[c]{@{}l@{}}Dual branch-BWCE+BHP\\ w/o cycle update stategy\end{tabular} & HPM                 & \begin{tabular}[c]{@{}c@{}}85.65\\ (0.48)\end{tabular}          & \begin{tabular}[c]{@{}c@{}}73.48\\ (0.48)\end{tabular}          & \begin{tabular}[c]{@{}c@{}}83.00\\ (1.60)\end{tabular}          & \begin{tabular}[c]{@{}c@{}}77.40\\ (0.60)\end{tabular}          & \begin{tabular}[c]{@{}c@{}}96.65\\ (0.15)\end{tabular}          \\\hline
Dual branch-BWCE+BHP                                                                    & 2 proxies per-class & \begin{tabular}[c]{@{}c@{}}85.52\\ (0.03)\end{tabular}          & \begin{tabular}[c]{@{}c@{}}74.03\\ (0.28)\end{tabular}          & \begin{tabular}[c]{@{}c@{}}81.46\\ (0.12)\end{tabular}          & \begin{tabular}[c]{@{}c@{}}77.22\\ (0.13)\end{tabular}          & \begin{tabular}[c]{@{}c@{}}96.53\\ (0.03)\end{tabular}          \\\hline
Dual branch-BWCE+BHP                                                                    & 3 proxies per-class & \begin{tabular}[c]{@{}c@{}}85.36\\ (0.09)\end{tabular}          & \begin{tabular}[c]{@{}c@{}}73.49\\ (0.10)\end{tabular}          & \begin{tabular}[c]{@{}c@{}}83.00\\ (0.33)\end{tabular}          & \begin{tabular}[c]{@{}c@{}}77.53\\ (0.20)\end{tabular}          & \begin{tabular}[c]{@{}c@{}}96.74\\ (0.02)\end{tabular}          \\\hline
Dual branch-BWCE+BHP                                                                    & 4 proxies per-class & \begin{tabular}[c]{@{}c@{}}85.79\\ (0.03)\end{tabular}          & \begin{tabular}[c]{@{}c@{}}74.09\\ (0.62)\end{tabular}          & \begin{tabular}[c]{@{}c@{}}82.03\\ (0.66)\end{tabular}          & \begin{tabular}[c]{@{}c@{}}77.42\\ (0.56)\end{tabular}          & \begin{tabular}[c]{@{}c@{}}96.53\\ (0.03)\end{tabular}          \\\hline
Dual branch-BWCE+BHP                                                                    & HPM                 & \textbf{\begin{tabular}[c]{@{}c@{}}86.11\\ (0.16)\end{tabular}} & \textbf{\begin{tabular}[c]{@{}c@{}}76.57\\ (0.94)\end{tabular}} & \textbf{\begin{tabular}[c]{@{}c@{}}83.22\\ (0.06)\end{tabular}} & \textbf{\begin{tabular}[c]{@{}c@{}}79.46\\ (0.58)\end{tabular}} & \textbf{\begin{tabular}[c]{@{}c@{}}96.78\\ (0.09)\end{tabular}} \\ \hline\hline
\end{tabular}
\end{table}

\end{document}